\documentclass{sig-alternate-05-2015}
\usepackage{color}

\usepackage{algorithm}
\usepackage{algpseudocode}
\usepackage{pifont}
\usepackage{gensymb}
\usepackage{pbox}
 \usepackage{array,multirow,graphicx,lmodern, tabularx}
\begin{document}

\setcopyright{acmcopyright}



\conferenceinfo{KDD '16 Workshop: Data Science for Food, Energy and Water}{August 14, 2016, San Francisco, CA, USA}


%

\title{A Bayesian Network approach to County-Level {\ttlit Corn Yield Prediction} using historical data and expert knowledge
}
%
%
%
%
%

\numberofauthors{6} 
%
\author{
%
%
\alignauthor
Vikas Chawla \thanks{The presenting author is an early researcher who wishes to be considered for the travel grant option.\vspace{1mm}}\\
       \affaddr{Department of Computer Science}\\
       \affaddr{Iowa State University}\\
       \affaddr{Ames, IA 50011}\\
       \email{vchawla@iastate.edu}
\alignauthor
Hsiang Sing Naik\\
       \affaddr{Department of Mechanical Engineering}\\
       \affaddr{Iowa State University}\\
       \affaddr{Ames, IA 50011}\\
       \email{hsnaik@iastate.edu}
\alignauthor
Adedotun Akintayo\\
       \affaddr{Department of Mechanical Engineering}\\
       \affaddr{Iowa State University}\\
       \affaddr{Ames, IA 50011}\\
       \email{akintayo@iastate.edu}
\and  
\alignauthor
Dermot Hayes\\
       \affaddr{Department of Economics}\\
       \affaddr{Iowa State University}\\
       \affaddr{Ames, IA 50011}\\
       \email{dhayes@iastate.edu}
\alignauthor
Patrick Schnable\\
       \affaddr{Plant Sciences Institute}\\
       \affaddr{Iowa State University}\\
       \affaddr{Ames, IA 50011}\\
       \email{schnable@iastate.edu}
\alignauthor
Baskar Ganapathysubramanian\\
       \affaddr{Department of Mechanical Engineering}\\
       \affaddr{Iowa State University}\\
       \affaddr{Ames, IA 50011}\\
       \email{baskarg@iastate.edu}
\and
\alignauthor
Soumik Sarkar \thanks{Corresponding author}\\
       \affaddr{Department of Mechanical Engineering}\\
       \affaddr{Iowa State University}\\
       \affaddr{Ames, IA 50011}\\
       \email{soumiks@iastate.edu}
}


\maketitle
\begin{abstract}
Crop yield forecasting is the methodology of predicting crop yields prior to harvest. The availability of accurate yield prediction frameworks have enormous implications from multiple standpoints, including impact on the crop commodity futures markets, formulation of agricultural policy, as well as crop insurance rating. The focus of this work is to construct a corn yield predictor at the county scale. Corn yield (forecasting) depends on a complex, interconnected set of variables that include economic, agricultural, management and meteorological factors. Conventional forecasting is either knowledge-based computer programs (that simulate plant-weather-soil-management interactions) coupled with targeted surveys or statistical model based. The former is limited by the need for painstaking calibration, while the latter is limited to univariate analysis or similar simplifying assumptions that fail to capture the complex interdependencies affecting yield. In this paper, we propose a data-driven approach that is \lq gray box\rq \hspace{1mm} i.e. that seamlessly utilizes expert knowledge in constructing a statistical network model for corn yield forecasting. Our multivariate gray box model is developed on \textit{Bayesian network analysis} to build a \textit{Directed Acyclic Graph (DAG)} between predictors and yield. Starting from a complete graph connecting various carefully chosen variables and yield, expert knowledge is used to prune or strengthen edges connecting variables. Subsequently the structure (connectivity and edge weights) of the DAG that maximizes the likelihood of observing the training data is identified via optimization. We curated an extensive set of historical data ($1948-2012$) for each of the $99$ counties in Iowa as data to train the model. We discuss preliminary results, and specifically focus on (a) the structure of the learned network and how it corroborates with known trends, and (b) how partial information still produces reasonable predictions (predictions with gappy data), and show that incorporating the missing information improves predictions.
\end{abstract}

%
%


%
%

%
%



\section{Introduction and Related Work} \label{sec:intro}
Crop yield forecasting is the methodology of predicting crop yields (at various scales: from farms to counties, to countries and to global scale) prior to harvest. Accurate crop yield predictions have enormous implications from multiple standpoints. These include: the impact on the crop commodity futures markets, timely interventions for crop management, unraveling genetic-environment interactions (GxE) for plant breeding, and appropriate policy decisions in both developing countries where food shortages remain a threat and in US where improved yield forecasting can improve targeting of conservation funding from major federal programs such as the Conservation Reserve Program.
\begin{figure*}[htbp]
\centering
 \includegraphics[width=1.0\textwidth,trim={0 0 10 0}]{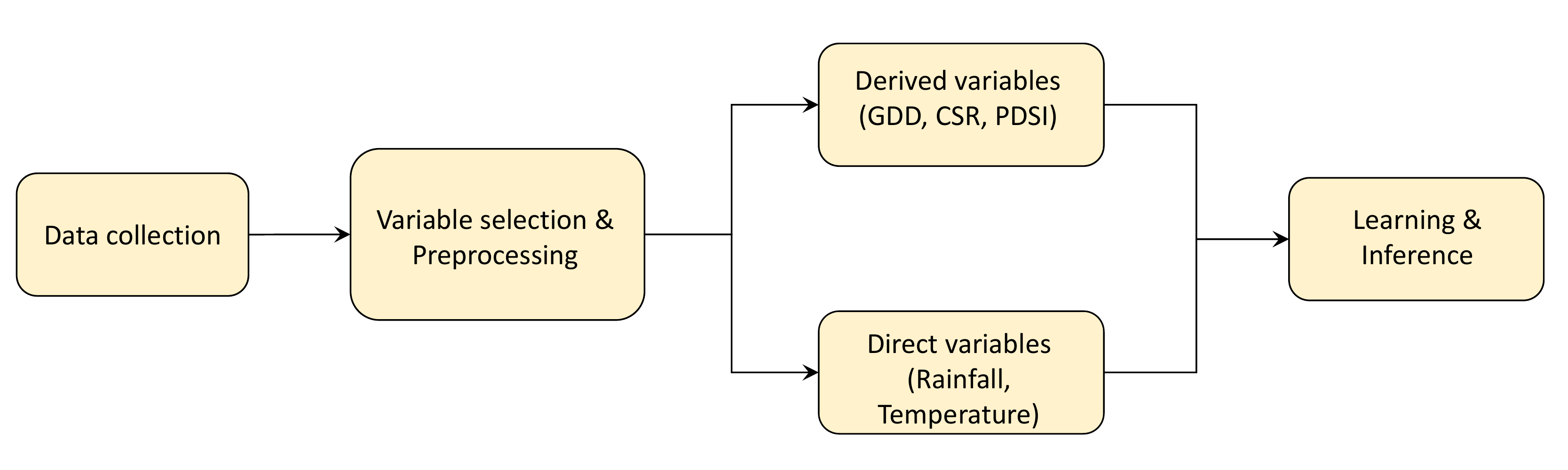}
\caption{Schematic of the yield prediction workflow}
\label{fig:flowchart}
\end{figure*}

The United States is the largest producer of corn in the world. Exports of corn alone account for approximately 10-20\% of annual revenue in the trade market. In the United States corn is grown nationwide, but production is mainly concentrated in the heartland region which includes Iowa and Illinois. Government and insurance companies have established a compensation system that insures farmers to support them against natural causes that have adverse effects on yield, but their premium rates are reported to be too high~\cite{SWD14, TY11}. On the other hand, any fluctuations in the corn futures market can have a debilitating impact on farmers. Therefore, the U.S. Department of Agriculture (USDA) invests an enormous amount of time and financial resources to making periodic county level yield predictions. This helps keep market participants equally informed about events that influence cash and futures prices for major commodities in an effort to prevent market failure due to non-participation by uninformed groups. The intellectual foundation behind this effort, described in a Nobel Prize winning paper on ``The Market for Lemons'' by George Akerlof, is that markets will fail if one set of participants have more information than other participants. Recent developments in the way agricultural information is collected and shared suggests that companies and big data firms may now be able to beat the USDA at this activity leading to detrimental asymmetric markets. A publicly available high quality yield prediction tool will enable the producers to make informed decisions thereby ensuring a symmetrical market. This is the motivation for the current work.

Conventional crop forecasting relies on a combination of knowledge-based computer programs (that simulate plant-weather-soil-management interactions) along with soil and environment data and targeted surveys or is based on statistical black-box approaches. The former is limited by the need for painstaking calibration, while the latter is limited to univariate analysis or similar simplifying assumptions that fail to capture the complex interdependencies affecting yield~\cite{2, 3, KQ14}. In this paper, we tread a middle ground between so-called \lq black-box\rq \hspace{1mm} and \lq white-box\rq \hspace{1mm} approaches. We present a  novel, knowledge-based statistical forecasting  approach to predict county-wide corn yield in the state of Iowa.  Our multivariate \lq gray box\rq \hspace{1mm} model is based on \textit{Bayesian Networks} and is utilized to build a \textit{Directed Acyclic Graph (DAG)} between predictors and yield. This  mathematical construct is implemented in a freely available reasoning engine for graphical models, SMILE, along with its graphical user interface (GUI), GeNIe~\cite{MD99}. We curated an extensive set of historical data ($1948-2012$) for each of the $99$ counties in Iowa for use as training data for the model. This historical weather data ($1948-2012$) was tediously collected from several public sources such as the National Agricultural Statistics Service (NASS), and included weather, topographic/soil, and some management traits. We utilize expert knowledge for variable selection and for graph pruning, and present promising initial results. Results include yield forecasts for all counties and a discussion of prediction accuracy;  an illustration of how prediction is possible with incomplete information, and the possibility of a probabilistic graphical model to perform what-if scenario analysis.

\section{Methodology} \label{sec:method}
Corn yield depends on a complex set of economical, meteorological, agricultural and financial inputs. These inputs are most likely interdependent. Formulating a \lq mechanistic model \rq (i.e. \lq knowledge--based\rq \hspace{1mm} models, or those based on mathematically defined equation(s)) relating inputs with output seems (currently) intractable. However, there is a large amount of historical data across geographical regions available that can be used to make future yield prediction. The availability of a corpus of historical data along with advances in \lq gray box\rq \hspace{1mm} machine learning models motivate us to utilize this approach to yield prediction. Probabilistic graphical models (PGM's) are an example of such \lq gray box\rq \hspace{1mm} machine learning (ML) models that are helpful in capturing conditional and causal dependencies; spatially, temporally and spatial-temporally. PGM's naturally allow for incorporation of expert knowledge and derive scientific understanding form the learnt models. Inference process in such Bayesian networks can be used for prediction and also for exploring \emph{What-if} scenarios; thus allowing us to perform inference on specific explanatory variables and observing changes in trends. PGM's are also scalable and are capable of handling large data sets. More attractively, they are capable of working with missing and conflicting data, and can inherently handle uncertainty. We outline a schematic of our workflow in Figure.~\ref{fig:flowchart}.

\subsection{Data Collection and Curation}
The focus of the data collection was getting a historical record of various explanatory variables and county yields for the 99 counties of the state of Iowa. We divided this task into two stages: 1) Collecting raw data from a variety of sources, and 2) Data curation, to organize the collected raw data in a form that is compatible with the machine learning framework, GeNIe. The weather data is taken from the Global Historical Climatology Network (GHCN) database which is hosted by the National Climatic Data Center (NCDC). We chose to utilize weather data from the months of May \ - September. This choice simply tracks the corn growing season over most of the corn belt region across Iowa. We assume that explanatory variables of time periods outside the growing season have negligible effect on end-of-season yield harvest. Relaxation of such assumptions will be explored in the future. The county scale soil data is taken from the Soil Survey Geographic (SSURGO) database that is hosted by the USDA. The collected data was then post-processed into expert knowledge derived variables -- specifically, aggregating daily temperatures into monthly averages, converting daily temperature into Growing Degree Days (GDD), an agronomic means of keeping track of heat. Further details of the data set, along with descriptions of each derived variable are provided later in the text. Data is curated for $99$ counties over a time period of 64 years ($1948$ to $2012$). The total dataset collected has an approximate size of $500$ MB and is stored in comma-separated values (CSV) file format. Our preliminary results are based on a subset of this data. We focus on a recent six year duration of $2005$--$2010$, with $5$ years used as training data, and the data from $2010$ used as testing data to explore the model's predictive capability.

\subsection{Variable Selection and Preprocessing} \label{sub:varsel}
Variable selection is critical to the construction of a viable yield predictor. We utilize expert knowledge (via agronomic arguments) to chose a subset of all possible inputs affecting yield in order to construct our probabilistic graphical model. We detail each variable and the rationale for the specific choice next.

\subsubsection{Growing Degree Days (GDD) or Heat Units} \label{sub:gdd}
The growth rate of corn is highly dependant on temperature. Ideal temperature conditions for robust growth is between a minimum temperature of 50$^{\circ}$F (10$^{\circ}$C), upto an optimum temperature of 86$^{\circ}$F (30$^{\circ}$C). Growth rates have been observed to decline if temperatures do not fall within this range. The Growing Degree Days (GDD) is an agronomic variable that represents the relationship between temperature and growth rate~\cite{6}. GDD is a heuristic tool in phenology that measures heat accumulation to predict development rates.  GDD is given by  \[GDD = (T_{max} + T_{min})/2 - T_{base} \]
where,
\begin{itemize}
  \item $T_{max}$ is the maximum daily temperature or equal to 86$^{\circ}$F (30$^{\circ}$C) when temperature exceed beyond 86$^{\circ}$F (30$^{\circ}$C).
  \item $T_{min}$ is the minimum daily temperature or equal to 50$^{\circ}$F (10$^{\circ}$C) when temperature falls below 50$^{\circ}$F (10$^{\circ}$C).
  \item $T_{base}$ is the base temperature required to trigger the optimum growth.
\end{itemize}
An additional motivation to choose this variable is the possibility of integrating seed type as an explanatory variable in the future. Seed companies typically report hybrid maturity in days and in terms of GDD. These reports are linked to the expected number of days necessary to reach enough GDD (about 2700 to 3100 GDD to reach \emph{R6} (physiological maturity)) to complete growth and development. For example, the commonly used 111 day hybrid requires approximately 111 days to attain enough GDD for harvest maturity.

\subsubsection{Palmer Drought Severity Index (PDSI)} \label{sub:pdsi}
Drought has a critical impact on farming and yield. The Palmer Drought Severity Index (PDSI) measures the availability of moisture after precipitation and recent temperature changes. It is based on the supply and demand concept of the water balance equation and considers multiple meteorological parameters (including water content in the soil, rate of evapotranspiration, soil recharge and moisture loss from the surface layer). The PDSI has also been used to perform spatial, and temporal correlations analysis~\cite{7}. The PDSI~\textsuperscript{\ref{footnote-label}} takes a value of 0 to indicate the normal conditions, negative values indicate drought severity and positive values indicate wetland or flooded conditions.

\subsubsection{Corn Suitability Rating (CSR2)} \label{sub:csr}
Soil type impacts productivity potential, and combined with weather conditions, is considered a dominant factor influencing yield. Corn Suitability Rating (CSR2) is an integrated measure based on soil mineral content, topographic features like slope gradient and slope length that indicate the suitability of the soil to grow corn. CSR2 ratings~\footnote{\label{footnote-label} In Figure.~\ref{fig:netwk} and~\ref{fig:learnstruc}, \textquotedblleft \emph{DI\_Avg}\textquotedblright \hspace{0.5mm} represent annual average PDSI values~\cite{7} and \textquotedblleft \emph{Soil\_WA}\textquotedblright \hspace{0.5mm} represent weighted average CSR2 ratings~\cite{CSR2} for each of the 99 counties in Iowa.} varies minimally over time and usually range from 5 \ - 100, with higher ratings correlating to better growing conditions.

\subsubsection{Rainfall} \label{sub:rainfall}
Precipitation is a factor that strongly affects yield. During the growing season, moisture requirements have to be met by rainfall, or through water held within the soil prior to growing season. High yield harvest within the corn belt region of the US has been due to the amount of precipitation available (>45cm) throughout the growing season. The demand for water utilization increases when the corn plant nears the tasseling stage, usually around mid-July, extending to mid-August. Note that both inadequate as well as over abundant rainfall reduce corn yields.

\subsubsection{Data Discretization} \label{sub:ddiscrete}
Before any network or structure is learnt, the available dataset is first categorized into a set of bins. This data transformation is necessary since our model is based on discrete Bayesian networks where modeling of the relationship is required in a parsimonious manner. The goal is to retain the underlying relationship between the variables while reducing the effects of external disturbances that may distort the relationship.  We chose to use a hierarchical discretization~\cite{RK92} over uniform width or uniform count. This enables automatic determination of  the optimal number of bins and their widths, given the multivariate distribution of the variables.


\subsubsection{Incorporating Background Knowledge}
\begin{figure*}[htbp]
\vskip -0.30in
\centering
 \includegraphics[width=1.0\textwidth,trim={0 0 10 0}]{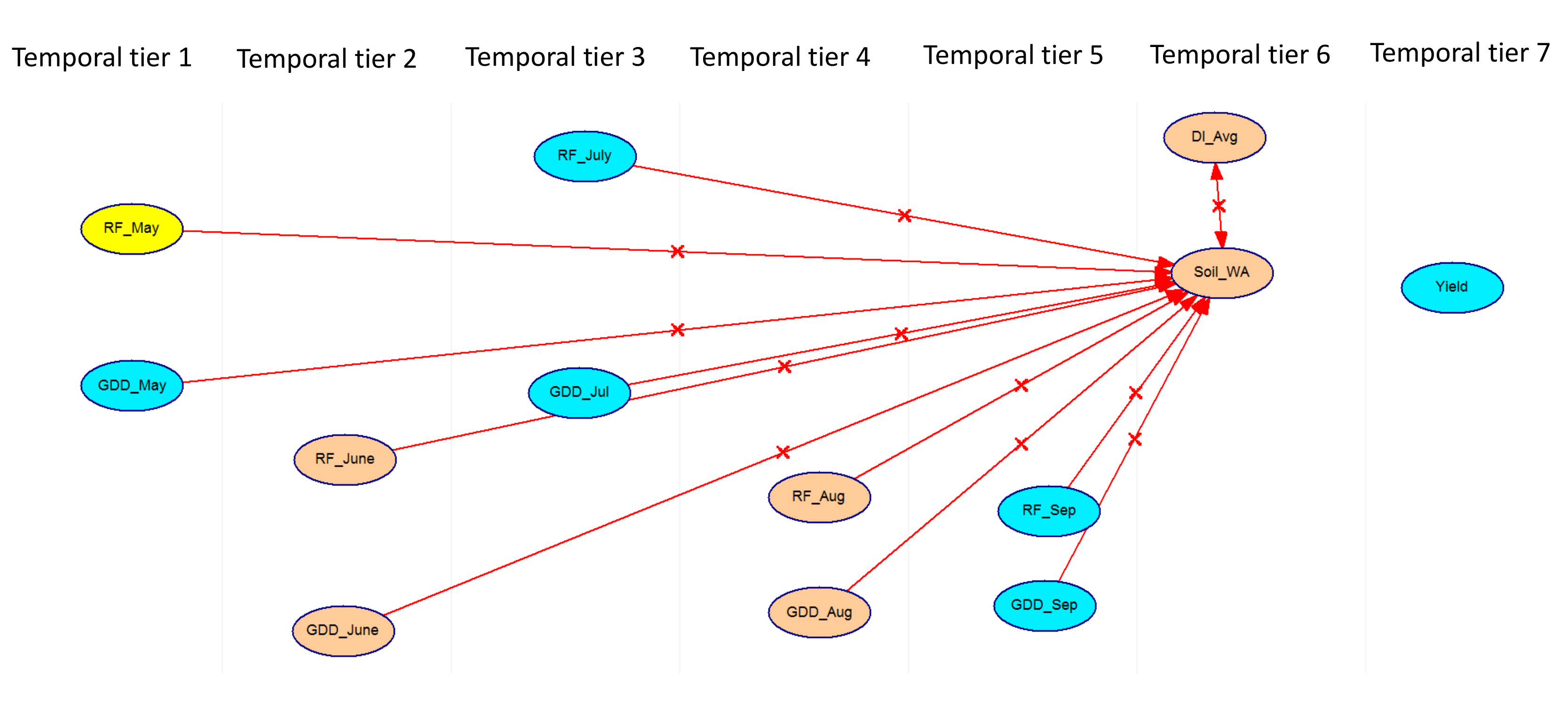}
\caption{Tiering and partial enforcing of Bayesian Network Structure with Prior Background Knowledge}
\label{fig:netwk}
\end{figure*}

The ability to include domain knowledge in the construction of a model is one of the strong points for the probabilistic graphical modeling technique. This allows domain experts to provide quality input regarding known correlations between variables, as connections (or edges) in the graph. Domain expertise enabled us to specify a strong link between rainfall and yield. This approach also allowed domain experts to forbid connections between specific variables (either through intuition or where such lack-of-correlation has been previously shown). This is extremely useful when working with temporally-sensitive data, allowing one to forbid connections from future observations to past observations. It is also important for the scalability of the structure learning stage. Furthermore, it allows the sorting of variables in temporal tiers, which also forbids future to past connections. Figure.~\ref{fig:netwk} displays the implemented background knowledge for our model.


\begin{figure*}[htbp]
\vskip -0.5in
\centering
 \includegraphics[width=1.0\textwidth,trim={0 0 10 0}]{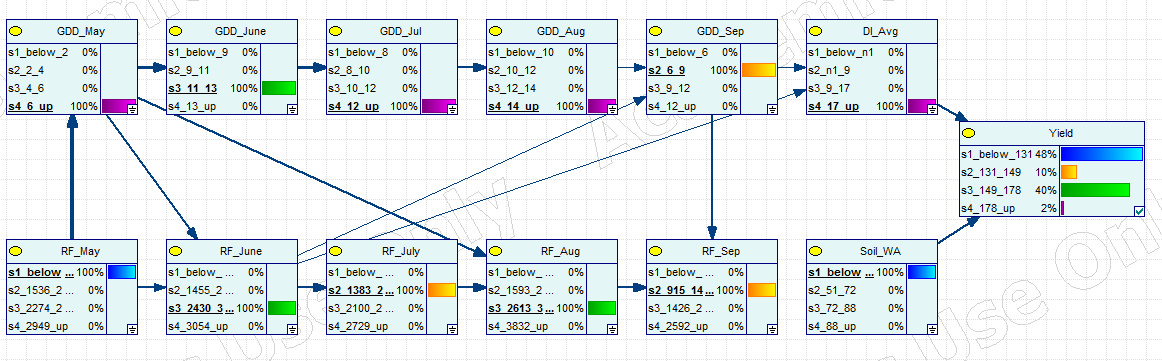}
 \caption{Illustration of the learnt Bayesian Network Structure based on Background knowledge}
\label{fig:learnstruc}
\end{figure*}

\subsection{Learning and Inference} \label{sub:landinf}
Learning and inference are the two main steps associated with graphical models such as Bayesian networks. Learning refers to training the probabilistic graphical model with the training data and the inference step involves decision making using the trained model and testing data/evidence. Learning/training involves identifying the structure (the DAG, or the edges of the graph) and learning the parameters (the edge weights), i.e., the conditional probability densities. The goal is to identify the structure and the associated parameters that best explain the given training data.

Given a Markovian set of variables $\textbf{x} := (x_1, \cdots, x_l)$, a \textit{DAG}, $\mathcal{G} = (\mathcal{V}, \mathcal{E})$ and a $\textbf{P}_\theta$ where $\mathcal{V}$ describes the set of nodes in the model, $\mathcal{E}$ gives the edges connecting nodes. $\textbf{P}_\theta(\textbf{x})$ represents the joint probability distribution factored on the variables given their parent nodes and $\theta$ describes the parameters learnt in the factoring process. More detailed descriptions of such models are available in vast amount of literature~\cite{ADA13, DCMHM14}. Mathematically, the aim of the learning task is to determine the optimal set of $(\mathcal{V}, \mathcal{E})$ as well as $\theta$ that describes the relationship embedded in the factors and the class variable (in this case, yield). Finding the optimal Bayesian network structure is an NP-hard problem, but efficient algorithms are available that often yield near optimal solutions~\cite{krishnamurthy2014scalable}. Bayesian networks support learning in supervised as well as in unsupervised settings, and thereby can be used with both labeled and unlabeled data sets (such as for knowledge discovery).

In this study, after discretizing the training data, we learned a network structure (Directed Acyclic Graph) that maximizes the likelihood of observing the training data. As mentioned earlier, finding such a \textit{DAG} is an NP-hard problem, hence we used efficient heuristics to approximate the underlying structure. Also, we sought expert knowledge in order to make the structure search more efficient. This knowledge elicitation helps the algorithm to streamline its connectivity search since we forbid some unreasonable links and force links where we have information related to conditional dependencies among variables. It is important to penalize dense structures as they typically lead to over-parameterization and hence, over-fitting (bias-variance tradeoff). To address this tradeoff, we track the Bayesian Information Criterion (BIC) to drive our search for the best DAG. A set of scoring functions such as minimum description length, MDL, Bayesian-Dirichlet functions and their variations~\cite{NL10} for learning DAG structures were introduced in~\cite{BM14}. Figure.~\ref{fig:learnstruc} shows the Bayesian Network structure that was learned via GeNIe toolbox on the so far curated training dataset. Note, the thickness of an edge between a pair of nodes reflects the degree of statistical dependency between those nodes i.e., strength of influence~\cite{krishnamurthy2014scalable}.

Inference pertains to finding probabilistic answers to user specified queries. For example, a user may seek the joint distribution of a subset of random variables given the observed values of other independent subsets of the random variables. Since Bayesian networks only encode node-wise conditional probabilities, finding answers to such queries is not straightforward. However, efficient algorithms exist that allow one to find the exact answer to an arbitrary query using a secondary structure (such as junction tree) and a message-passing architecture~\cite{krishnamurthy2014scalable}.

GeNIe has in-built support for various learning algorithms. In this paper, we employed the Bayesian search algorithm to train the model. It is a general purpose graph structure learning algorithm that makes use of the Bayesian search procedure to explore the full space of graphs, $\mathcal{G}$. In this case, the posterior probability tables are filled out using expectation maximization algorithm,  
\[ \arg\max_\mathcal{G} P(\mathcal{G}|D)\]
given the data, D. The aim of the algorithm is to run partial search over Markov equivalence class of the data instead of directly searching over the full \textit{DAG}s space to reduce the computation time. Note that a Markov equivalence class~\cite{DCMHM14} is a subset graph class that contains both directed and undirected edges, i.e., it is a set containing all the DAGs that are Markov equivalent to each other.




In the implementation of Bayesian search in GeNIe, we added background knowledge by forbidding $20$ edges.  The tiering edges ($i->tier$) that associates nodes with particular tier in the $7$--tier model is shown in Figure.~\ref{fig:netwk}. 







\subsubsection{Expected yield prediction} \label{sub:accuracy}
Given that the model structure and the parameters of a $DAG$ have been learnt, it is necessary to make inferences on the model by getting forecast of yield in terms of expected yield. Accuracy of the model is tested based on the available evidence to calculate the difference in the predicted and actual yield. Given, historical values of yield \emph{Y} (in bu/ac), we define \emph{$\hat{Y}$} as the expected yield prediction provided that we have computed the posterior distribution $P(b_n)$ during the inference process where $b_n$ is the $n^{th}$ bin signifying a certain range of yield. With this setup, we have\\
\[  \emph{$\hat{Y}$}= \sum_{n} P(b_n) \cdot E(Y|b_n) \]
where,
\begin{itemize}
  \item $n \in \{1,\cdots, 4\}$ denotes the discrete bin for the yield variable.
  \item $P(b_n)$ denotes the probability of yield being in the range marked by bin $b_n$.
  \item $E(Y|b_n)$ represents the expected yield in the bin $b_n$ computed based on the training data.
\end{itemize}


\section{Results and discussion} \label{sec:result}

In this section, initial results are presented for the Bayesian network based county level yield prediction approach.
We used $2005$--$2009$ data in this study and the data set was divided into a training and testing set. While $75$\% of the data was used for learning the Bayes Net structure and parameters, the remaining $25$\% was used to provide an in-sample validation for the model. The validation set is used to determine the effectiveness of the model; to estimate its accuracy and the confidence level; to analyze performance with incomplete and complete evidence and to examine various `what-if' scenarios as described below.


\subsection{Yield prediction}~\label{sub:yielpred}
\begin{table}[h]
\begin{tabular}{ |p{1.8cm}|p{1cm}|p{1.1cm}|p{1.1cm}|p{1.4cm}|}
\hline
& \multicolumn{4}{|c|}{Predicted yield (in Bu/ac)} \\
\hline
True yield (in Bu/ac)& 0--131 &131--149&149--178&178--Above \\
\hline
0--131 & 6 & 0 & 0 & 0\\ \hline
131--149 & 4   & 11 & 0 & 0 \\ \hline
149--178 & 0 & 1 & 14 & 7\\ \hline
178--Above& 2 & 0 & 6 & 46 \\
\hline
\end{tabular}
\caption{Confusion Matrix with four yield level classes}
\label{table:3}
\end{table}








The effectiveness of our model is described using a confusion matrix shown in the Table~\ref{table:3}. 
It shows the overall capability of the model to correctly categorize predicted yields in the validation set into the appropriate bins, i.e., yield prediction ranges.
While most of the data is in the diagonal (i.e., correct prediction), some of the estimated yields fall into the wrong bins. However, in most cases the miss-predictions fall into neighboring bins which suggests small errors. Moreover, this current study uses an incomplete set of explanatory variables and we are currently expanding the set of variables to utilize cumulative effects of temperature and localized effects of rainfall.

\begin{table}[h]
\centering
\begin{tabular}{|p{2cm}|p{1.5cm}|p{1.5cm}|p{1.5cm}| }
\hline
\textbf{County} & \textbf{Actual Yield {Bu/ac}} & \textbf{Predicted Yield (Bu/ac)} & \textbf{Difference (\%)} \\ \hline
Shelby         & 171.6          & 171.71                    & 0.06              \\ \hline
Bremer  & 174.6            & 174.39                   & 0.12              \\ \hline
Palo Alto         & 174              & 174.39       & 0.22              \\ \hline
Calhoun   & 173.3          & 174.39                   & 0.63               \\ \hline
\end{tabular}
\caption{Difference between Predicted and Actual Yield at a county level}
\label{table:1}
\end{table}

Table~\ref{table:1} displays sample results of expected yield (as described in~\ref{sub:accuracy}) obtained from the model. 
The model was used to predict yield in all 99 counties of Iowa in 2010 and overall, predicted yield for 70 out of the 99 counties had an accuracy of 80\% or more. 
This illustrates the yield prediction potential of a Bayesian Network model with reasonable explanatory variables and domain knowledge embedding. However, this is still an on-going effort and we are working to include more key variables and domain knowledge for better prediction accuracy.  

\subsection{Prediction with partial and complete evidences} \label{sub:part}

\begin{table*}[htbp]
\centering
\begin{tabular}{|p{3.5cm}|p{2.5cm}|p{2.5cm}|p{1.5cm}|p{1.5cm}|p{1.5cm}| }
\hline
Evidences &Time Period&\textbf{County} & \textbf{Actual Yield (Bu/ac)} & \textbf{Predicted Yield (Bu/ac)} & \textbf{Difference (\%)} \\ \hline
{GDD \& RF} & May--June & Polk & 139.40   & 167.91   & 30 \\ \hline
{GDD \& RF} & May--July & Polk & 139.40   & 167.91   & 30 \\ \hline
{GDD \& RF} & May--August & Polk & 139.40   & 167.91   & 30 \\ \hline
{GDD \& RF} & May--September & Polk & 139.40   & 165.55  & 29 \\ \hline
{GDD, RF, PDSI \& CSR2} & May--September & Polk & 139.40   & 140.88   & 2 \\ \hline
\end{tabular}
\caption{Table showing the effects of gradual addition of evidence on selected counties' yield prediction accuracy}
\label{tab:comvsincom}
\end{table*}

The ultimate goal of this research is a publicly available high quality yield prediction tool that will enable the producers to make informed decisions. From this perspective, the tool needs to start predicting yield estimates from early part of the season and aim to improve the prediction as season moves forward and more observations are used as evidence. In this context, Bayesian network is an ideal inference framework as it can function with missing variables/data unlike many other approaches such as standard regression. We investigated the yield prediction performance in the absence of complete evidence--that is, before the end of the growing season, where information on future weather conditions is unavailable. Note, in such a scenario, a model can still use future weather predictions which can potentially help such a tool positively. However, we did not consider availability of any such predicted weather conditions in this study. In this case study, initial (incomplete) evidence includes only the growing degree days (GDD) and rainfall (RF) for the months of May--June. Then as the season progresses, we added evidence from months of July, August and September respectively. Furthermore, we added key variables such as PDSI and CSR2 at the final stage to examine the improvement in yield prediction performance.
\\The effect of incomplete evidence for Polk county is shown in the Table.~\ref{tab:comvsincom}.
With initial limited evidence, the model is capable of providing a reasonable estimate of yield and as expected, performance improves with added evidence and finally with complete evidence\footnote{Note that the term complete evidence in this case is based on the data available for this study which is far from being exhaustive.}, the computed yield comes very close to the actual yield (lagging the actual by only $\approx 1 (Bu/ac)$). This is an illustration of how a Bayesian Network based tool can be leveraged seamlessly for continuous yield prediction throughout the growing season.



\subsection{ What-If Scenarios} \label{sub:whatif}
\begin{figure}[!htb]
        \centering
        \includegraphics[scale=0.6]{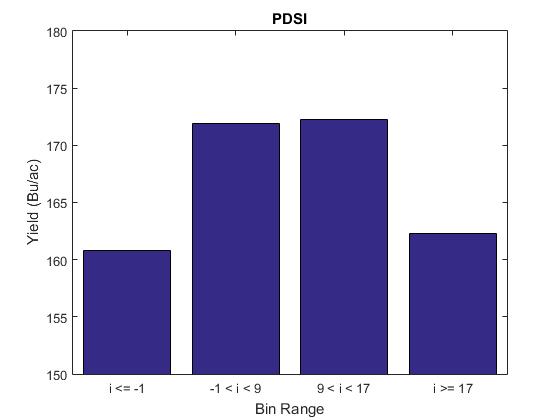}
        \caption{Histogram of inference on expected yield of PDSI}
        \label{fig:infres}
\end{figure}
\begin{figure}[!htb]
        \centering
        \includegraphics[scale=0.6]{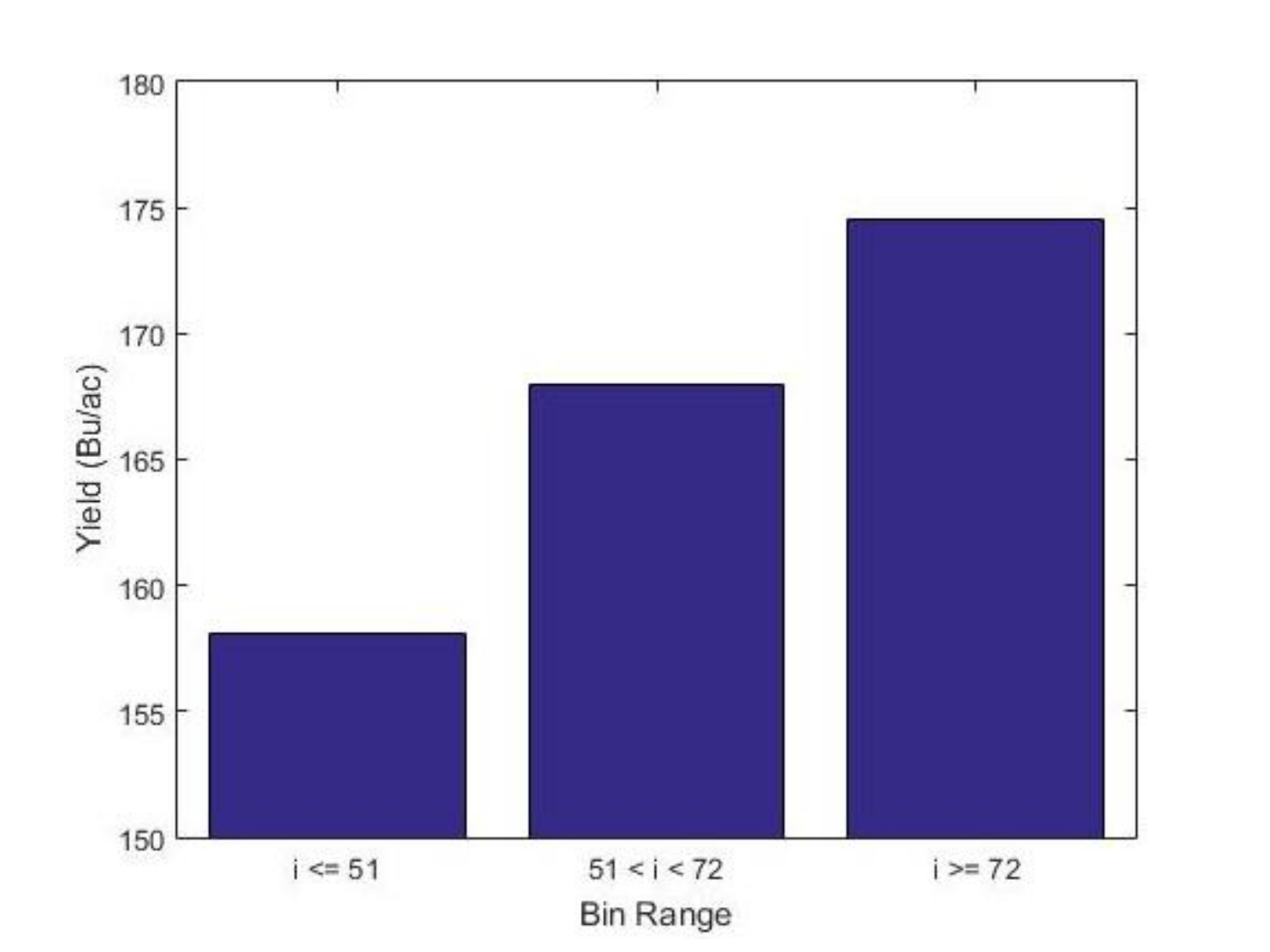}
        \caption{Histogram of inference on expected yield of CSR2}
        \label{fig:infcsr}
\end{figure}
\begin{figure}[!htb]
        \centering
        \includegraphics[scale=0.6]{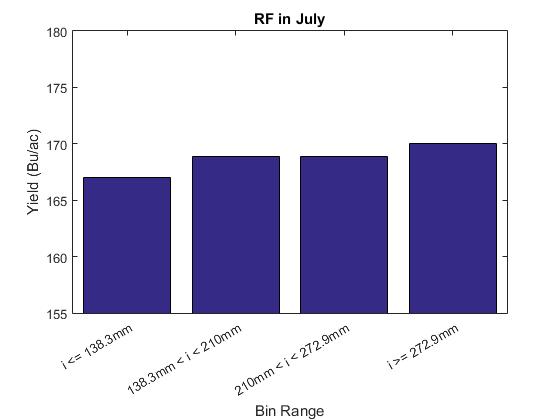}
        \caption{Histogram of inference on expected yield of rainfall in july}
        \label{fig:infrfjuly}
\end{figure}

Farmers and plant scientists are extremely interested in learning key driving variables and parameters that affect yield. 
In this context, a probabilistic graphical model such as Bayesian Network can be an effective tool to understand the impact of different variables (e.g., weather) on a certain target variable (e.g., yield). Such an inference exercise is called simulation of `what-if' scenarios and a few examples are provided below:

It is known that a host of the climatic factors lead to drop in expected corn yields at extreme conditions. A good example to support that is the effect that PDSI, described in subsection~\ref{sub:pdsi}, has on the estimated yield. 
Figure~\ref{fig:infres} shows the result of a `what-if' scenario simulation where bins 1 and 4 for PDSI lead to lower yield compared to bins 2 and 3. Note, bins 1 and 4 suggest highly negative or highly positive PDSI values which indicate extreme drought or extreme wet conditions respectively whereas bins 2 and 3 contain PDSI values that are around zero which indicate a close to ideal condition. Thus the Bayes Net inference result conforms with the scientific knowledge that extreme dry or extreme wet conditions are both bad for corn yield.


In addition to PDSI, the effect of CSR2 on yield is examined and the result is shown in Figure.~\ref{fig:infcsr}. There is a reasonable positive correlation between the CSR2 values and expected yield confirming the domain knowledge of farmers and plant scientists.

Another example is shown in Figure.~\ref{fig:infrfjuly} where increased rainfall in July tends to help corn production slightly.
In summary, a Bayesian Network model is not only useful for yield prediction but also effective for understanding various causal effects (unlike different black box models) that can enhance the scientific knowledge in this domain.

\section{Summary, Conclusions and Fu-\\ture Work}

In this paper, we demonstrated a Bayesian Network approach in order to predict county-wide yield in the corn belt state of Iowa, primarily utilizing historical weather data. Apart from the yield prediction capability with incomplete and complete evidence, key advantages of such an approach include ability to incorporate domain knowledge, enhance scientific understanding via `what-if' scenario simulation and naturally provide a prediction confidence. In the case study presented here, the model performed reasonably well based on its validation accuracy. Example `what-if' scenarios involving PDSI, CSR2 and rainfall in July show effectiveness of this approach in enhancing scientific understanding. We also demonstrated the capability of yield prediction based on incomplete and complete evidence which makes it a useful tool for continuous yield prediction throughout the season. While the main future goal of this research is to be able to accurately predict yield within 5 Bu/ac of the actual yield in every county, many other technical aspects are being pursued as well such as (i) incorporation of cumulative weather variables, (ii) handling different time-scales of different explanatory variables and (iii) establishing a model adaptation mechanism along with climate change patterns.


\section{Acknowledgments}
Vikas Chawla and Baskar Ganapathysubramanian thank ISU PSI for support through the PSI faculty fellow. All authors thank the ISU PIIR DDSI funding for partial support.  %
\bibliographystyle{abbrv}
\bibliography{sigproc}  
%
%

\end{document}